\documentclass[11pt,a4paper]{article}
\usepackage[hyperref]{acl2017}
\usepackage{times}
\usepackage{latexsym}
\usepackage{graphicx}
\usepackage{float}
\usepackage{amsmath}
\usepackage{amssymb}
\usepackage{multirow}
\usepackage{booktabs}
\usepackage{url}
\usepackage{array}

\aclfinalcopy 
\def\aclpaperid{18} 

\title{Attention-over-Attention Neural Networks for Reading Comprehension}

\author{Yiming Cui$^\dag$, Zhipeng Chen$^\dag$, Si Wei$^\dag$, Shijin Wang$^\dag$, Ting Liu$^\ddag$ \and Guoping Hu$^\dag$\\
{$^\dag$Joint Laboratory of HIT and iFLYTEK, iFLYTEK Research, Beijing, China}\\
{$^\ddag$Research Center for Social Computing and Information Retrieval,}\\
{Harbin Institute of Technology, Harbin, China}\\
{$^\dag$\tt\{ymcui,zpchen,siwei,sjwang3,gphu\}@iflytek.com}\\  
{$^\ddag$\tt tliu@ir.hit.edu.cn}\\
}

\date{}
\begin{document}
\maketitle

\begin{abstract}
Cloze-style reading comprehension is a representative problem in mining relationship between document and query.
In this paper, we present a simple but novel model called {\em attention-over-attention} reader for better solving cloze-style reading comprehension task.
The proposed model aims to place another attention mechanism over the document-level attention and induces ``attended attention'' for final answer predictions.
One advantage of our model is that it is simpler than related works while giving excellent performance.
In addition to the primary model, we also propose an N-best re-ranking strategy to double check the validity of the candidates and further improve the performance.
Experimental results show that the proposed methods significantly outperform various state-of-the-art systems by a large margin in public datasets, such as CNN and Children's Book Test.
\end{abstract}

\section{Introduction}\label{introduction}
To read and comprehend the human languages are challenging tasks for the machines, which requires that the understanding of natural languages and the ability to do reasoning over various clues. 
Reading comprehension is a general problem in the real world, which aims to read and comprehend a given article or context, and answer the questions based on it. 
Recently, the cloze-style reading comprehension problem has become a popular task in the community.
The cloze-style query \cite{taylor-etal-1953} is a problem that to fill in an appropriate word in the given sentences while taking the context information into account.

To teach the machine to do cloze-style reading comprehensions, large-scale training data is necessary for learning relationships between the given document and query.
To create large-scale training data for neural networks, \newcite{hermann-etal-2015} released the CNN/Daily Mail news dataset, where the document is formed by the news articles and the queries are extracted from the summary of the news. 
\newcite{hill-etal-2015} released the Children's Book Test dataset afterwards, where the training samples are generated from consecutive 20 sentences from books, and the query is formed by 21st sentence. 
Following these datasets, a vast variety of neural network approaches have been proposed \citep{kadlec-etal-2016,cui-etal-2016,chen-etal-2016,dhingra-etal-2016,sordoni-etal-2016,trischler-etal-2016,seo-etal-2016,xiong-etal-2016}, and most of them stem from the attention-based neural network \cite{bahdanau-etal-2014}, which has become a stereotype in most of the NLP tasks and is well-known by its capability of learning the ``importance'' distribution over the inputs.

In this paper, we present a novel neural network architecture, called {\em attention-over-attention} model. 
As we can understand the meaning literally, our model aims to place another attention mechanism over the existing document-level attention.
Unlike the previous works, that are using heuristic merging functions \cite{cui-etal-2016}, or setting various pre-defined non-trainable terms \cite{trischler-etal-2016}, our model could automatically generate an ``attended attention'' over various document-level attentions, and make a mutual look not only from {\em query-to-document} but also {\em document-to-query}, which will benefit from the interactive information. 

To sum up, the main contributions of our work are listed as follows.
\begin{itemize}
	\item To our knowledge, this is the first time that the mechanism of nesting another attention over the existing attentions is proposed, i.e. {\em attention-over-attention} mechanism.
	\item Unlike the previous works on introducing complex architectures or many non-trainable hyper-parameters to the model, our model is much more simple but outperforms various state-of-the-art systems by a large margin.
	\item We also propose an N-best re-ranking strategy to re-score the candidates in various aspects and further improve the performance.
\end{itemize}

The following of the paper will be organized as follows.
In Section \ref{rc-task}, we will give a brief introduction to the cloze-style reading comprehension task as well as related public datasets. 
Then the proposed attention-over-attention reader will be presented in detail in Section \ref{nn-for-rc} and N-best re-ranking strategy in Section \ref{reranking}.
The experimental results and analysis will be given in Section \ref{experiments} and Section \ref{analysis}.
Related work will be discussed in Section \ref{related-work}.
Finally, we will give a conclusion of this paper and envisions on future work.

\section{Cloze-style Reading Comprehension}\label{rc-task}
In this section, we will give a brief introduction to the cloze-style reading comprehension task at the beginning. 
And then, several existing public datasets will be described in detail.

\subsection{Task Description}
Formally, a general Cloze-style reading comprehension problem can be illustrated as a triple:
\begin{equation}
\nonumber \langle \mathcal D, \mathcal Q, \mathcal A \rangle
\end{equation}
The triple consists of a document  $\mathcal D$, a query $\mathcal Q$ and the answer to the query $\mathcal A$. 
Note that the answer is usually a {\em single} word in the document, which requires the human to exploit context information in both document and query.
The type of the answer word varies from predicting a preposition given a fixed collocation to identifying a named entity from a factual illustration.

\subsection{Existing Public Datasets}
Large-scale training data is essential for training neural networks.
Several public datasets for the cloze-style reading comprehension has been released. 
Here, we introduce two representative and widely-used datasets.

\subsubsection*{$\bullet$~~ CNN / Daily Mail}
\newcite{hermann-etal-2015} have firstly published two datasets: CNN and Daily Mail news data \footnote{The pre-processed CNN and Daily Mail datasets are available at \url{http://cs.nyu.edu/~kcho/DMQA/}}.
They construct these datasets with web-crawled CNN and Daily Mail news data. 
One of the characteristics of these datasets is that the news article is often associated with a summary. 
So they first regard the main body of the news article as the {\em Document}, and the {\em Query} is formed by the summary of the article, where one entity word is replaced by a special placeholder to indicate the missing word. 
The replaced entity word will be the {\em Answer} of the {\em Query}.  
Apart from releasing the dataset, they also proposed a methodology that anonymizes the named entity tokens in the data, and these tokens are also re-shuffle in each sample. The motivation is that the news articles are containing limited named entities, which are usually celebrities, and the world knowledge can be learned from the dataset. So this methodology aims to exploit general relationships between anonymized named entities within a single document rather than the common knowledge.
The following research on these datasets showed that the entity word anonymization is not as effective as expected \citep{chen-etal-2016}.

\subsubsection*{$\bullet$~~ Children's Book Test}
There was also a dataset called the Children's Book Test (CBTest) released by \newcite{hill-etal-2015}, which is built on the children's book story through Project Gutenberg \footnote{The CBTest datasets are available at \url{http://www.thespermwhale.com/jaseweston/babi/CBTest.tgz}}. 
Different from the CNN/Daily Mail datasets, there is no summary available in the children's book. So they proposed another way to extract query from the original data.
The document is composed of 20 consecutive sentences in the story, and the 21st sentence is regarded as the query, where one word is blanked with a special placeholder.
In the CBTest datasets, there are four types of sub-datasets available which are classified by the part-of-speech and named entity tag of the answer word, containing Named Entities (NE), Common Nouns (CN), Verbs and Prepositions. 
In their studies, they have found that the answering of verbs and prepositions are relatively less dependent on the content of document, and the humans can even do preposition blank-filling without the presence of the document.
The studies shown by \newcite{hill-etal-2015}, answering verbs and prepositions are less dependent with the presence of document. Thus, most of the related works are focusing on solving NE and CN types.

\section{Attention-over-Attention Reader}\label{nn-for-rc}

\begin{figure*}[htbp]
  \centering
  \includegraphics[width=1\textwidth]{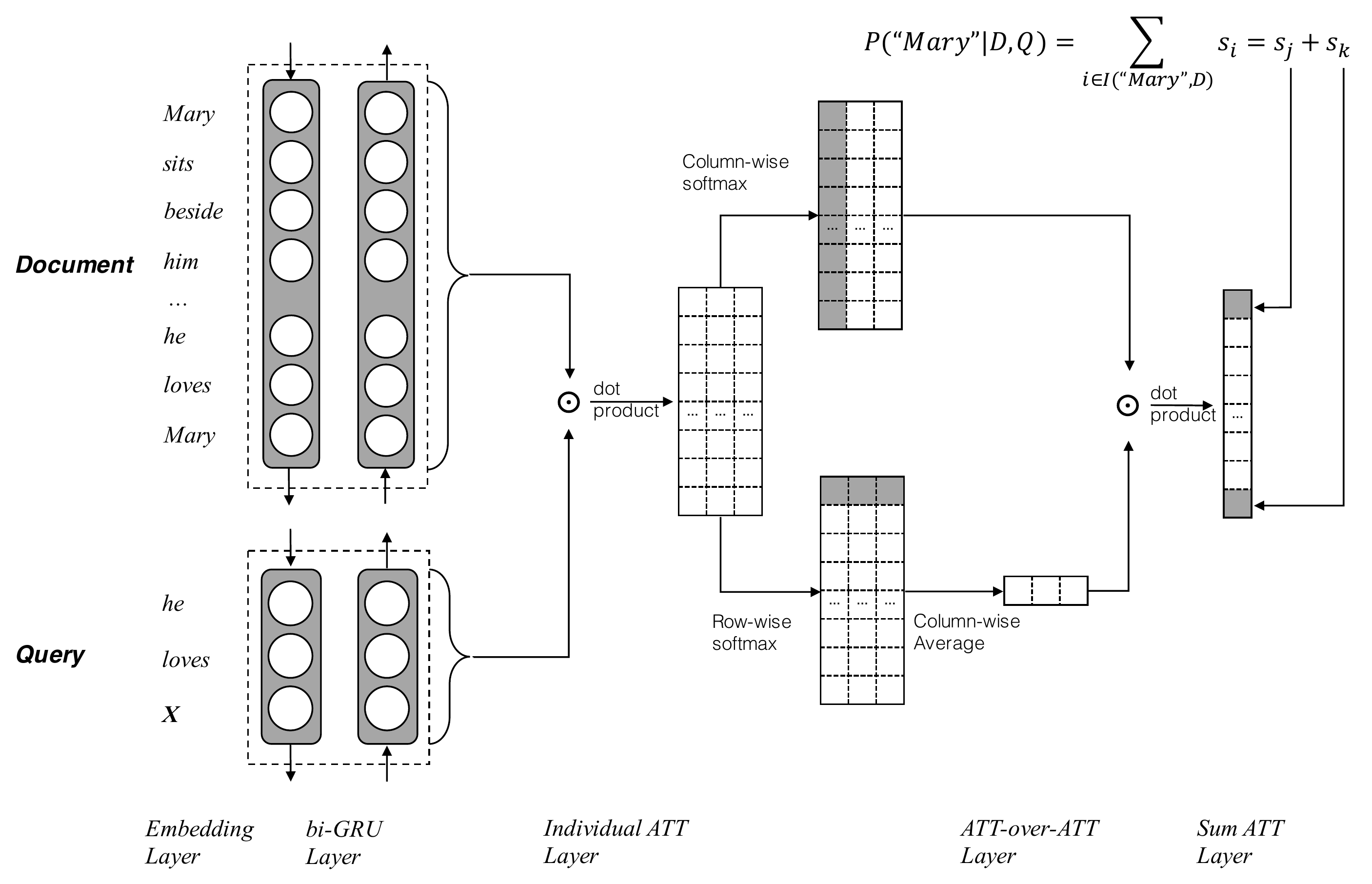}
  \caption{\label{nn-arch} Neural network architecture of the proposed Attention-over-Attention Reader (AoA Reader).}
\end{figure*}

In this section, we will give a detailed introduction to the proposed Attention-over-Attention Reader (AoA Reader).
Our model is primarily motivated by Kadlec et al., \shortcite{kadlec-etal-2016}, which aims to directly estimate the answer from the document-level attention instead of calculating blended representations of the document. 
As previous studies by \newcite{cui-etal-2016} showed that the further investigation of query representation is necessary, and it should be paid more attention to utilizing the information of query.
In this paper, we propose a novel work that placing another attention over the primary attentions, to indicate the ``importance'' of each attentions.

Now, we will give a formal description of our proposed model.
When a cloze-style training triple $\langle \mathcal D, \mathcal Q, \mathcal A \rangle$ is given, the proposed model will be constructed in the following steps.

\subsubsection*{$\bullet$~~ Contextual Embedding}
We first transform every word in the document $\mathcal D$ and query $\mathcal Q$ into one-hot representations and then convert them into continuous representations with a shared embedding matrix $W_e$. 
By sharing word embedding, both the document and query can participate in the learning of embedding and both of them will benefit from this mechanism. 
After that, we use two bi-directional RNNs to get contextual representations of the document and query individually, where the representation of each word is formed by concatenating the forward and backward hidden states.
After making a trade-off between model performance and training complexity, we choose the Gated Recurrent Unit (GRU) \cite{cho-etal-2014} as recurrent unit implementation.
\begin{gather}
e(x) = W_e \cdot x,~where~~x\in \mathcal D , \mathcal Q \\
\overrightarrow{h_s(x)} =  \overrightarrow{GRU}(e(x)) \\
\overleftarrow{h_s(x)} = \overleftarrow{GRU}(e(x)) \\
h_s(x) = [\overrightarrow{h_s(x)}; \overleftarrow{h_s(x)}]
\end{gather}

We take $h_{doc}\in\mathbb{R}^{|\mathcal D|*2d}$ and $h_{query}\in\mathbb{R}^{|\mathcal Q|*2d}$ to denote the contextual representations of document and query, where $d$ is the dimension of GRU (one-way).

\subsubsection*{$\bullet$~~ Pair-wise Matching Score}
After obtaining the contextual embeddings of the document $h_{doc}$ and query $h_{query}$, we calculate a pair-wise matching matrix, which indicates the pair-wise matching degree of one document word and one query word. Formally, when given $i$th word of the document and $j$th word of query, we can compute a matching score by their dot product.
\begin{equation} M(i,j) = h_{doc}(i)^{T} \cdot h_{query}(j) \end{equation}

In this way, we can calculate every pair-wise matching score between each document and query word, forming a matrix $M\in\mathbb{R}^{|\mathcal D|*|\mathcal Q|}$, where the value of $i$th row and $j$th column is filled by $M(i,j)$.

\subsubsection*{$\bullet$~~ Individual Attentions}
After getting the pair-wise matching matrix $M$, we apply a column-wise softmax function to get probability distributions in each column, where each column is an individual document-level attention when considering a single query word. We denote $\alpha(t)\in\mathbb{R}^{|\mathcal D|}$ as the document-level attention regarding query word at time $t$, which can be seen as a {\em query-to-document} attention. 
\newcommand\D{\displaystyle}
\begin{gather}
\alpha(t) = softmax(M(1,t),...,M(|\mathcal D|,t)) \\
\alpha = [\alpha(1), \alpha(2), ..., \alpha(|\mathcal Q|)]
\end{gather}

\subsubsection*{$\bullet$~~ Attention-over-Attention}
Different from \newcite{cui-etal-2016}, instead of using naive heuristics (such as {\em summing} or {\em averaging}) to combine these individual attentions into a final attention, we introduce another attention mechanism to automatically decide the importance of each individual attention.

First, we calculate a reversed attention, that is, for every document word at time $t$, we calculate the ``importance'' distribution on the query, to indicate which query words are more important given a single document word.
We apply a row-wise softmax function to the pair-wise matching matrix $M$ to get query-level attentions. We denote $\beta(t)\in\mathbb{R}^{|\mathcal Q|}$ as the query-level attention regarding document word at time $t$, which can be seen as a {\em document-to-query} attention. 
\begin{equation} \beta(t) = softmax(M(t,1),...,M(t,|\mathcal Q|)) \end{equation}

So far, we have obtained both {\em query-to-document} attention $\alpha$ and  {\em document-to-query} attention $\beta$. 
Our motivation is to exploit mutual information between the document and query. 
However, most of the previous works are only relying on {\em query-to-document} attention, that is, only calculate one document-level attention when considering the whole query. 

Then we average all the $\beta(t)$ to get an averaged query-level attention $\beta$. Note that, we do not apply another softmax to the $\beta$, because averaging individual attentions do not break the normalizing condition.
\begin{equation} \beta = \frac{1}{n}\sum\limits_{t=1}^{|\mathcal D|}\beta(t) \end{equation}

Finally, we calculate dot product of $\alpha$ and $\beta$ to get the ``attended document-level attention'' $s\in\mathbb{R}^{|\mathcal D|}$, i.e. the {\em attention-over-attention} mechanism. 
Intuitively, this operation is calculating a weighted sum of each individual document-level attention $\alpha(t)$ when looking at query word at time $t$.
In this way, the contributions by each query word can be learned explicitly, and the final decision (document-level attention) is made through the voted result by the importance of each query word.
\begin{equation} s = \alpha^{T} \beta  \end{equation}

        \begin{table*}[htbp]
        \begin{center}
        \begin{tabular}{lrrrrrrrrr}
        \toprule
        & \multicolumn{3}{c}{CNN News} & \multicolumn{3}{c}{CBT NE} & \multicolumn{3}{c}{CBT CN} \\
        & Train & Valid & Test & Train & Valid & Test & Train & Valid & Test \\
        \midrule
        \# Query & 380,298 & 3,924 & 3,198 & 108,719 & 2,000 & 2,500 & 120,769 & 2,000 & 2,500 \\
        Max \# candidates & 527 & 187 & 396 & 10 & 10 & 10 & 10 & 10 & 10 \\
        Avg \# candidates & 26 & 26 & 25 & 10 & 10 & 10 & 10 & 10 & 10 \\
        Avg \# tokens & 762 & 763 & 716 & 433 & 412 & 424 & 470 & 448 & 461 \\
        Vocabulary & \multicolumn{3}{c}{118,497} & \multicolumn{3}{c}{53,063} & \multicolumn{3}{c}{53,185}\\
        \bottomrule
        \end{tabular}
        \end{center}
        \caption{\label{cbt-stats} Statistics of cloze-style reading comprehension datasets: CNN news and CBTest NE / CN.}
        \end{table*}

\subsubsection*{$\bullet$~~ Final Predictions}
Following \newcite{kadlec-etal-2016}, we use {\em sum attention} mechanism to get aggregated results. Note that the final output should be reflected in the vocabulary space $V$, rather than document-level attention $|\mathcal D|$, which will make a significant difference in the performance, though \newcite{kadlec-etal-2016} did not illustrate this clearly. 
\begin{equation} P(w|\mathcal D, \mathcal Q) = \sum_{i \in I(w,\mathcal D)} s_i  ,~~w \in V \end{equation}

where $I(w,\mathcal D)$ indicate the positions that word $w$ appears in the document $\mathcal D$. 
As the training objectives, we seek to maximize the log-likelihood of the correct answer.
\begin{equation} \mathcal{L} = \sum_{i} \log(p(x))~~, x\in\mathcal{A}\end{equation}

The proposed neural network architecture is depicted in Figure \ref{nn-arch}.
Note that, as our model mainly adds limited steps of calculations to the AS Reader \cite{kadlec-etal-2016} and does not employ any additional weights, the computational complexity is similar to the AS Reader.

\section{N-best Re-ranking Strategy}\label{reranking}

Intuitively, when we do cloze-style reading comprehensions, we often refill the candidate into the blank of the query to double-check its appropriateness, fluency and grammar to see if the candidate we choose is the most suitable one. If we do find some problems in the candidate we choose, we will choose the second possible candidate and do some checking again.

To mimic the process of double-checking, we propose to use N-best re-ranking strategy after generating answers from our neural networks. 
The procedure can be illustrated as follows.

\subsubsection*{$\bullet$~~ N-best Decoding}
Instead of only picking the candidate that has the highest possibility as answer, we can also extract follow-up candidates in the decoding process, which forms an N-best list.

\subsubsection*{$\bullet$~~ Refill Candidate into Query}
As a characteristic of the cloze-style problem, each candidate can be refilled into the blank of the query to form a complete sentence.
This allows us to check the candidate according to its context.

\subsubsection*{$\bullet$~~ Feature Scoring}
The candidate sentences can be scored in many aspects.
In this paper, we exploit three features to score the N-best list.

\begin{itemize}
  \item Global N-gram LM: This is a fundamental metric in scoring sentence, which aims to evaluate its fluency. This model is trained on the document part of training data.
  \item Local N-gram LM: Different from global LM, the local LM aims to explore the information with the given document, so the statistics are obtained from the test-time document. It should be noted that the local LM is trained sample-by-sample, it is not trained on the entire test set, which is not legal in the real test case. This model is useful when there are many unknown words in the test sample.
  \item Word-class  LM: Similar to global LM, the word-class LM is also trained on the document part of training data, but the words are converted to its word class ID. The word class can be obtained by using clustering methods. In this paper, we simply utilized the {\em mkcls} tool for generating 1000 word classes \cite{och-1999}.
\end{itemize}

\subsubsection*{$\bullet$~~ Weight Tuning}
To tune the weights among these features, we adopt the K-best MIRA algorithm \cite{cherry-foster:2012:NAACL-HLT} to automatically optimize the weights on the validation set, which is widely used in statistical machine translation tuning procedure.

\subsubsection*{$\bullet$~~ Re-scoring and Re-ranking}
After getting the weights of each feature, we calculate the weighted sum of each feature in the N-best sentences and then choose the candidate that has the lowest cost as the final answer.

        \begin{table*}[h]
        \begin{center}
        \begin{tabular}{lcccccc}
        \toprule
        & \multicolumn{2}{p{2cm}}{\centering CNN News} & \multicolumn{2}{p{2cm}}{\centering CBTest NE} & \multicolumn{2}{p{2cm}}{\centering CBTest CN}\\
        & Valid & Test & Valid & Test & Valid & Test\\
        \midrule
        Deep LSTM Reader \cite{hermann-etal-2015} & 55.0 & 57.0 & - & - & - & - \\
        Attentive Reader \cite{hermann-etal-2015} & 61.6 & 63.0 & - & - & - & - \\
        Human (context+query) \cite{hill-etal-2015} & - & - & - & {\em 81.6} & - & {\em 81.6} \\
        MemNN (window + self-sup.) \cite{hill-etal-2015} & 63.4\qquad & 66.8 & 70.4 & 66.6 & 64.2 & 63.0 \\ 
        AS Reader \cite{kadlec-etal-2016} & 68.6 & 69.5 & 73.8 & 68.6 & 68.8 & 63.4 \\
        CAS Reader \cite{cui-etal-2016} & 68.2 & 70.0 & 74.2 & 69.2 & 68.2 & 65.7 \\
        Stanford AR \cite{chen-etal-2016} & 72.4 & 72.4 & - & - & - & - \\
        GA Reader \cite{dhingra-etal-2016} & 73.0 & 73.8 & 74.9 & 69.0 & 69.0 & 63.9 \\
        Iterative Attention \cite{sordoni-etal-2016} & 72.6 & 73.3 & 75.2 & 68.6 & 72.1 & 69.2 \\        
        EpiReader \cite{trischler-etal-2016} & {\bf 73.4} & 74.0 & 75.3 & 69.7 & 71.5 & 67.4 \\
        \hline
        {\bf AoA Reader} & 73.1 & {\bf 74.4} & {\bf 77.8} & {\bf 72.0} & {\bf 72.2} & {\bf 69.4} \\
        {\bf AoA Reader + Reranking} & - & - & {\bf 79.6} & {\bf 74.0} & {\bf 75.7} & {\bf 73.1} \\
        \hline\hline
        MemNN (Ensemble) & 66.2 & 69.4 & - & - & - & - \\
        AS Reader (Ensemble) & 73.9 & 75.4 & 74.5 & 70.6 & 71.1 & 68.9 \\
        GA Reader (Ensemble) & 76.4 & 77.4 & 75.5 & 71.9 & 72.1 & 69.4 \\
        EpiReader (Ensemble) & - & - & 76.6 & 71.8 & 73.6 & 70.6 \\
        Iterative Attention (Ensemble) & 74.5 & 75.7 & 76.9 & 72.0 & 74.1 & {\bf 71.0} \\
        \hline
        {\bf AoA Reader (Ensemble)} & - & - & {\bf 78.9} & {\bf 74.5} & {\bf 74.7} &  70.8 \\
        {\bf AoA Reader (Ensemble + Reranking)} & - & - & {\bf 80.3} & {\bf 75.6} & {\bf 77.0} &  {\bf 74.1} \\
        \bottomrule
        \end{tabular}
        \end{center}
        \caption{\label{public-result} Results on the CNN news, CBTest NE and CN datasets. The best baseline results are depicted in italics, and the overall best results are in bold face. 
         }
        \end{table*}
        
\section{Experiments}\label{experiments}

\subsection{Experimental Setups} 
        
The general settings of our neural network model are listed below in detail.

\begin{itemize}
  \item Embedding Layer: The embedding weights are randomly initialized with the uniformed distribution in the interval $[-0.05,0.05]$. For regularization purpose, we adopted $l_2$-regularization to 0.0001 and dropout rate of 0.1 \cite{srivastava-etal-2014}. Also, it should be noted that we do not exploit any pre-trained embedding models.
  \item Hidden Layer: Internal weights of GRUs are initialized with random orthogonal matrices \cite{saxe2013exact}.
  \item Optimization: We adopted ADAM optimizer for weight updating \cite{kingma2014adam}, with an initial learning rate of 0.001. As the GRU units still suffer from the gradient exploding issues, we set the gradient clipping threshold to 5 \cite{pascanu-etal-2013}. We used batched training strategy of 32 samples.
\end{itemize}

Dimensions of embedding and hidden layer for each task are listed in Table \ref{dim-stats}.
In re-ranking step, we generate 5-best list from the baseline neural network model, as we did not observe a significant variance when changing the N-best list size.
All language model features are trained on the training proportion of each dataset, with 8-gram word-based setting and Kneser-Ney smoothing \cite{kneser-1995} trained by SRILM toolkit \cite{stolcke-2002}.
The results are reported with the best model, which is selected by the performance of validation set.
The ensemble model is made up of four best models, which are trained using different random seed.
Implementation is done with Theano \cite{theano2016} and Keras \cite{chollet2015keras}, and all models are trained on Tesla K40 GPU.

        \begin{table}[hbp]
        \begin{center}
        \begin{tabular}{lccc}
        \toprule
        & Embed. \# units & Hidden \# units\\
        \midrule
        CNN News & 384 & 256 \\
        CBTest NE & 384 & 384 \\
        CBTest CN & 384 & 256 \\
        \bottomrule
        \end{tabular}
        \end{center}
        \caption{\label{dim-stats} Embedding and hidden layer dimensions for each task.}
        \end{table}
        
\subsection{Overall Results}        

Our experiments are carried out on public datasets: CNN news datasets \cite{hermann-etal-2015} and CBTest NE/CN datasets \cite{hill-etal-2015}.
The statistics of these datasets are listed in Table \ref{cbt-stats}, and the experimental results are given in Table \ref{public-result}.

As we can see that, our AoA Reader outperforms state-of-the-art systems by a large margin, where 2.3\% and 2.0\% absolute improvements over EpiReader in CBTest NE and CN test sets, which demonstrate the effectiveness of our model.
Also by adding additional features in the re-ranking step, there is another significant boost  2.0\% to 3.7\% over AoA Reader in CBTest NE/CN test sets.
We have also found that our single model could stay on par with the previous best ensemble system, and even we have an absolute improvement of 0.9\% beyond the best ensemble model (Iterative Attention) in the CBTest NE validation set. 
When it comes to ensemble model, our AoA Reader also shows significant improvements over previous best ensemble models by a large margin and set up a new state-of-the-art system. 

To investigate the effectiveness of employing {\em attention-over-attention} mechanism, we also compared our model to CAS Reader, which used pre-defined merging heuristics, such as {\em sum} or {\em avg} etc.
Instead of using pre-defined merging heuristics, and letting the model explicitly learn the weights between individual attentions results in a significant boost in the performance, where 4.1\% and 3.7\% improvements can be made in CNN validation and test set against CAS Reader.

\subsection{Effectiveness of Re-ranking Strategy}        

As we have seen that the re-ranking approach is effective in cloze-style reading comprehension task, we will give a detailed ablations in this section to show the contributions by each feature.
To have a thorough investigation in the re-ranking step, we listed the detailed improvements while adding each feature mentioned in Section \ref{reranking}. 

From the results in Table \ref{rerank-cbt}, we found that the NE and CN category both benefit a lot from the re-ranking features, but the proportions are quite different.
Generally speaking, in NE category, the performance is mainly boosted by the $LM_{local}$ feature.
However, on the contrary, the CN category benefits from $LM_{global}$ and $LM_{wc}$ rather than the $LM_{local}$.
           
Also, we listed the weights of each feature in Table \ref{weights-cbt}. The $LM_{global}$ and $LM_{wc}$ are all trained by training set, which can be seen as {\em Global Feature}. However, the $LM_{local}$ is only trained within the respective document part of test sample, which can be seen as {\em Local Feature}. 
\begin{equation}
\eta = \frac{LM_{global} + LM_{wc}}{LM_{local}}
\end{equation}

We calculated the ratio between the global and local features and found that the NE category is much more dependent on local features than CN category. Because it is much more likely to meet a new named entity than a common noun in the test phase, so adding the local LM provides much more information than that of common noun. However, on the contrary, answering common noun requires less local information, which can be learned in the training data relatively.

        \begin{table}[tp]
        \begin{center}
        \begin{tabular}{lcccc}
        \toprule
        & \multicolumn{2}{p{1.8cm}}{\centering CBTest NE} & \multicolumn{2}{p{1.8cm}}{\centering CBTest CN}\\
        & Valid & Test & Valid & Test\\
        \midrule
        AoA Reader & 77.8 & 72.0 & 72.2 & 69.4 \\
        +Global LM& 78.3 & 72.6 & 73.9 & 71.2 \\
        +Local LM& 79.4 & 73.8 & 74.7 & 71.7 \\
        +Word-class LM& 79.6 & 74.0 & 75.7 & 73.1 \\
        \bottomrule
        \end{tabular}
        \end{center}
        \caption{\label{rerank-cbt} Detailed results of 5-best re-ranking on CBTest NE/CN datasets. Each row includes all of the features from previous rows. $LM_{global}$ denotes the global LM, $LM_{local}$ denotes the local LM, $LM_{wc}$ denotes the word-class LM.
         }
        \end{table}
        
        \begin{table}[tp]
        \begin{center}
        \begin{tabular}{lcccc}
        \toprule
        & {\centering CBTest NE} & {\centering CBTest CN} \\
        \midrule
        NN & 0.64 & 0.20 \\
        Global LM & 0.16 & 0.10 \\
        Word-class LM & 0.04 & 0.39 \\
        Local LM & 0.16 & 0.31 \\
        \hline
        RATIO $\eta$ & 1.25 & 1.58 \\
        \bottomrule
        \end{tabular}
        \end{center}
        \caption{\label{weights-cbt} Weight of each feature in N-best re-ranking step. {\em NN} denotes the feature (probability) produced by baseline neural network model.
         }
        \end{table}

\section{Quantitative Analysis}\label{analysis}

\begin{figure}[ht]
  \centering
  \includegraphics[width=0.48\textwidth]{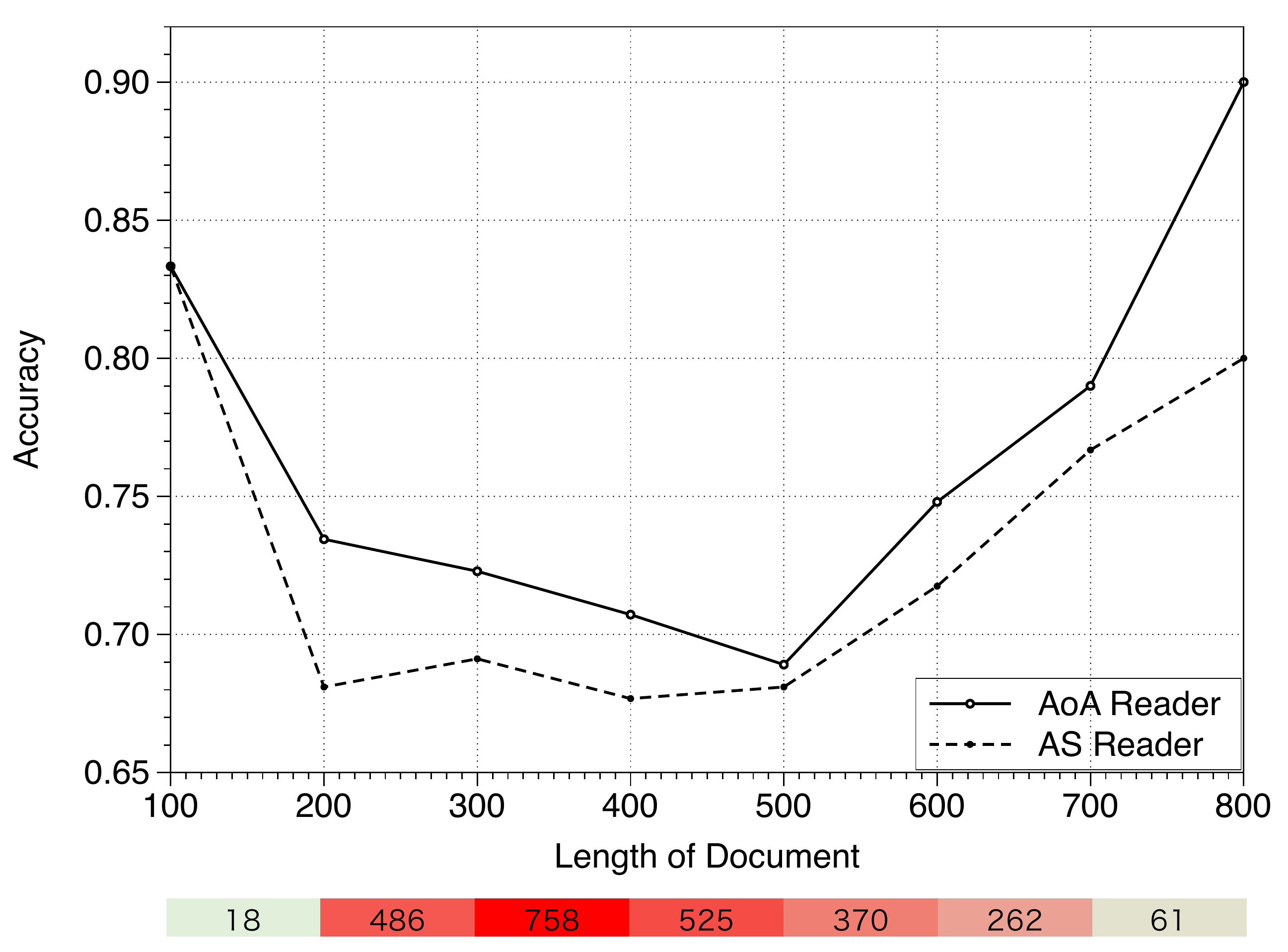}
  \caption{\label{length-acc} Test accuracy against the length of the document. The bar below the figure indicates the number of samples in each interval.}
\end{figure}

In this section, we will give a quantitative analysis to our AoA Reader. The following analyses are carried out on CBTest NE dataset.
First, we investigate the relations between the length of the document and corresponding accuracy. The result is depicted in Figure \ref{length-acc}. 

As we can see that the AoA Reader shows consistent improvements over AS Reader on the different length of the document.
Especially, when the length of document exceeds 700, the improvements become larger, indicating that the AoA Reader is more capable of handling long documents.

\begin{figure}[ht]
  \centering
  \includegraphics[width=0.48\textwidth]{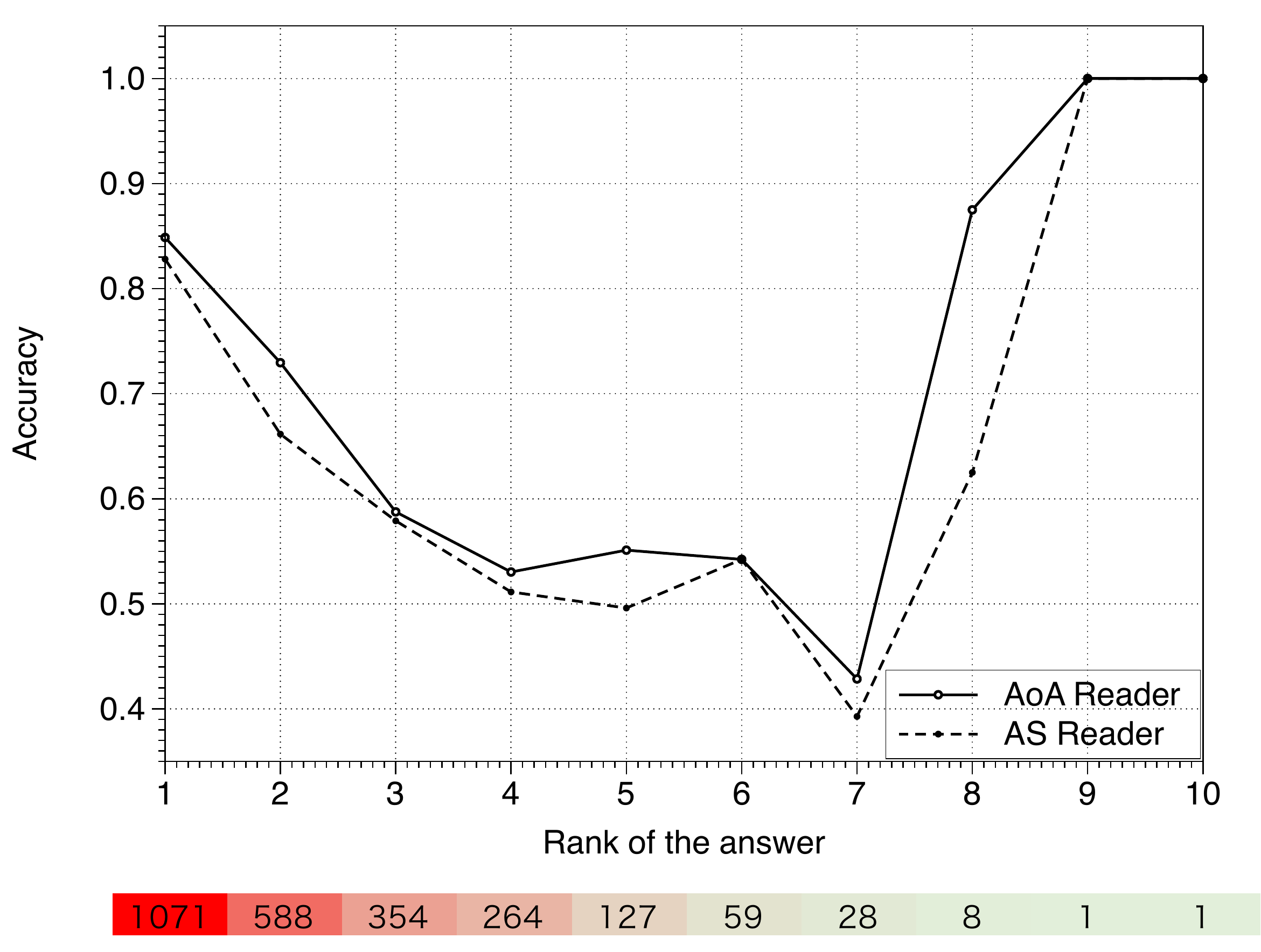}
  \caption{\label{rank-acc} Test accuracy against the frequency rank of the answer. The bar below the figure indicates the number of samples in each rank.}
\end{figure}

Furthermore, we also investigate if the model tends to choose a high-frequency candidate than a lower one, which is shown in Figure \ref{rank-acc}. 
Not surprisingly, we found that both models do a good job when the correct answer appears more frequent in the document than the other candidates.
This is because that the correct answer that has the highest frequency among the candidates takes up over 40\% of the test set (1071 out of 2500).
But interestingly we have also found that, when the frequency rank of correct answer exceeds 7 (less frequent among candidates), these models also give a relatively high performance.
Empirically, we think that these models tend to choose extreme cases in terms of candidate frequency (either too high or too low).
One possible reason is that it is hard for the model to choose a candidate that has a neutral frequency as the correct answer, because of its ambiguity (neutral choices are hard to made).

\section{Related Work}\label{related-work}

Cloze-style reading comprehension tasks have been widely investigated in recent studies. 
We will take a brief revisit to the related works.

\newcite{hermann-etal-2015} have proposed a method for obtaining large quantities of $\langle \mathcal D, \mathcal Q, \mathcal A \rangle$ triples through news articles and its summary. Along with the release of cloze-style reading comprehension dataset, they also proposed an attention-based neural network to handle this task. Experimental results showed that the proposed neural network is effective than traditional baselines.

\newcite{hill-etal-2015} released another dataset, which stems from the children's books. Different from \newcite{hermann-etal-2015}'s work, the document and query are all generated from the raw story without any summary, which is much more general than previous work. To handle the reading comprehension task, they proposed a window-based memory network, and self-supervision heuristics is also applied to learn hard-attention. 

Unlike previous works, that using blended representations of document and query to estimate the answer, \newcite{kadlec-etal-2016} proposed a simple model that directly pick the answer from the document, which is motivated by the Pointer Network \cite{vinyals-etal-2015}. A restriction of this model is that the answer should be a single word and appear in the document. Results on various public datasets showed that the proposed model is effective than previous works.

\citet{liu-etal-2016} proposed to exploit reading comprehension models to other tasks. They first applied the reading comprehension model into Chinese zero pronoun resolution task with automatically generated large-scale pseudo training data. The experimental results on OntoNotes 5.0 data showed that their method significantly outperforms various state-of-the-art systems.

Our work is primarily inspired by \newcite{cui-etal-2016} and \newcite{kadlec-etal-2016} , where the latter model is widely applied to many follow-up works \cite{sordoni-etal-2016,trischler-etal-2016,cui-etal-2016}. Unlike the CAS Reader \cite{cui-etal-2016}, we do not assume any heuristics to our model, such as using merge functions: $sum$, $avg$ etc. We used a mechanism called ``attention-over-attention'' to explicitly calculate the weights between different individual document-level attentions, and get the final attention by computing the weighted sum of them. Also, we find that our model is typically general and simple than the recently proposed model, and brings significant improvements over these cutting edge systems.

\section{Conclusion}\label{conclusion}

We present a novel neural architecture, called attention-over-attention reader, to tackle the cloze-style reading comprehension task. 
The proposed AoA Reader aims to compute the attentions not only for the document but also the query side, which will benefit from the mutual information. 
Then a weighted sum of attention is carried out to get an attended attention over the document for the final predictions. 
Among several public datasets, our model could give consistent and significant improvements over various state-of-the-art systems by a large margin. 

The future work will be carried out in the following aspects. 
We believe that our model is general and may apply to other tasks as well, so firstly we are going to fully investigate the usage of this architecture in other tasks. 
Also, we are interested to see that if the machine really ``comprehend'' our language by utilizing neural networks approaches, but not only serve as a ``document-level'' language model. In this context, we are planning to investigate the problems that need comprehensive reasoning over several sentences.

\section*{Acknowledgments}
We would like to thank all three anonymous reviewers for their thorough reviewing and providing thoughtful comments to improve our paper. This work was supported by the National 863 Leading Technology Research Project via grant 2015AA015409. 

\bibliography{acl2017}

\begin{thebibliography}{}
\expandafter\ifx\csname natexlab\endcsname\relax\def\natexlab#1{#1}\fi

\bibitem[{Bahdanau et~al.(2014)Bahdanau, Cho, and Bengio}]{bahdanau-etal-2014}
Dzmitry Bahdanau, Kyunghyun Cho, and Yoshua Bengio. 2014.
\newblock Neural machine translation by jointly learning to align and
  translate.
\newblock {\em arXiv preprint arXiv:1409.0473\/} .

\bibitem[{Chen et~al.(2016)Chen, Bolton, and Manning}]{chen-etal-2016}
Danqi Chen, Jason Bolton, and D.~Christopher Manning. 2016.
\newblock \href{https://doi.org/10.18653/v1/P16-1223}{A thorough examination of
  the cnn/daily mail reading comprehension task}.
\newblock In {\em Proceedings of the 54th Annual Meeting of the Association for
  Computational Linguistics (Volume 1: Long Papers)\/}. Association for
  Computational Linguistics, pages 2358--2367.
\newblock
  \href{https://doi.org/10.18653/v1/P16-1223}{https://doi.org/10.18653/v1/P16-1223}.

\bibitem[{Cherry and Foster(2012)}]{cherry-foster:2012:NAACL-HLT}
Colin Cherry and George Foster. 2012.
\newblock \href{http://www.aclweb.org/anthology/N12-1047}{Batch tuning
  strategies for statistical machine translation}.
\newblock In {\em Proceedings of the 2012 Conference of the North American
  Chapter of the Association for Computational Linguistics: Human Language
  Technologies\/}. Association for Computational Linguistics, Montr\'{e}al,
  Canada, pages 427--436.
\newblock
  \href{http://www.aclweb.org/anthology/N12-1047}{http://www.aclweb.org/anthology/N12-1047}.

\bibitem[{Cho et~al.(2014)Cho, van Merrienboer, Gulcehre, Bahdanau, Bougares,
  Schwenk, and Bengio}]{cho-etal-2014}
Kyunghyun Cho, Bart van Merrienboer, Caglar Gulcehre, Dzmitry Bahdanau, Fethi
  Bougares, Holger Schwenk, and Yoshua Bengio. 2014.
\newblock \href{http://aclweb.org/anthology/D14-1179}{Learning phrase
  representations using rnn encoder--decoder for statistical machine
  translation}.
\newblock In {\em Proceedings of the 2014 Conference on Empirical Methods in
  Natural Language Processing (EMNLP)\/}. Association for Computational
  Linguistics, pages 1724--1734.
\newblock
  \href{http://aclweb.org/anthology/D14-1179}{http://aclweb.org/anthology/D14-1179}.

\bibitem[{Chollet(2015)}]{chollet2015keras}
Fran\c{c}ois Chollet. 2015.
\newblock Keras.
\newblock \url{https://github.com/fchollet/keras}.

\bibitem[{Cui et~al.(2016)Cui, Liu, Chen, Wang, and Hu}]{cui-etal-2016}
Yiming Cui, Ting Liu, Zhipeng Chen, Shijin Wang, and Guoping Hu. 2016.
\newblock \href{http://aclweb.org/anthology/C16-1167}{Consensus attention-based
  neural networks for chinese reading comprehension}.
\newblock In {\em Proceedings of COLING 2016, the 26th International Conference
  on Computational Linguistics: Technical Papers\/}. The COLING 2016 Organizing
  Committee, pages 1777--1786.
\newblock
  \href{http://aclweb.org/anthology/C16-1167}{http://aclweb.org/anthology/C16-1167}.

\bibitem[{Dhingra et~al.(2016)Dhingra, Liu, Cohen, and
  Salakhutdinov}]{dhingra-etal-2016}
Bhuwan Dhingra, Hanxiao Liu, William~W Cohen, and Ruslan Salakhutdinov. 2016.
\newblock Gated-attention readers for text comprehension.
\newblock {\em arXiv preprint arXiv:1606.01549\/} .

\bibitem[{Hermann et~al.(2015)Hermann, Kocisky, Grefenstette, Espeholt, Kay,
  Suleyman, and Blunsom}]{hermann-etal-2015}
Karl~Moritz Hermann, Tomas Kocisky, Edward Grefenstette, Lasse Espeholt, Will
  Kay, Mustafa Suleyman, and Phil Blunsom. 2015.
\newblock Teaching machines to read and comprehend.
\newblock In {\em Advances in Neural Information Processing Systems\/}. pages
  1684--1692.

\bibitem[{Hill et~al.(2015)Hill, Bordes, Chopra, and Weston}]{hill-etal-2015}
Felix Hill, Antoine Bordes, Sumit Chopra, and Jason Weston. 2015.
\newblock The goldilocks principle: Reading children's books with explicit
  memory representations.
\newblock {\em arXiv preprint arXiv:1511.02301\/} .

\bibitem[{Josef~Och(1999)}]{och-1999}
Franz Josef~Och. 1999.
\newblock \href{http://aclweb.org/anthology/E99-1010}{An efficient method for
  determining bilingual word classes}.
\newblock In {\em Ninth Conference of the European Chapter of the Association
  for Computational Linguistics\/}.
\newblock
  \href{http://aclweb.org/anthology/E99-1010}{http://aclweb.org/anthology/E99-1010}.

\bibitem[{Kadlec et~al.(2016)Kadlec, Schmid, Bajgar, and
  Kleindienst}]{kadlec-etal-2016}
Rudolf Kadlec, Martin Schmid, Ond{\v{r}}ej Bajgar, and Jan Kleindienst. 2016.
\newblock \href{https://doi.org/10.18653/v1/P16-1086}{Text understanding with
  the attention sum reader network}.
\newblock In {\em Proceedings of the 54th Annual Meeting of the Association for
  Computational Linguistics (Volume 1: Long Papers)\/}. Association for
  Computational Linguistics, pages 908--918.
\newblock
  \href{https://doi.org/10.18653/v1/P16-1086}{https://doi.org/10.18653/v1/P16-1086}.

\bibitem[{Kingma and Ba(2014)}]{kingma2014adam}
Diederik Kingma and Jimmy Ba. 2014.
\newblock Adam: A method for stochastic optimization.
\newblock {\em arXiv preprint arXiv:1412.6980\/} .

\bibitem[{Kneser and Ney(1995)}]{kneser-1995}
Reinhard Kneser and Hermann Ney. 1995.
\newblock Improved backing-off for m-gram language modeling.
\newblock In {\em International Conference on Acoustics, Speech, and Signal
  Processing\/}. pages 181--184 vol.1.

\bibitem[{Liu et~al.(2016)Liu, Cui, Yin, Wang, Zhang, and Hu}]{liu-etal-2016}
Ting Liu, Yiming Cui, Qingyu Yin, Shijin Wang, Weinan Zhang, and Guoping Hu.
  2016.
\newblock Generating and exploiting large-scale pseudo training data for zero
  pronoun resolution.
\newblock {\em arXiv preprint arXiv:1606.01603\/} .

\bibitem[{Pascanu et~al.(2013)Pascanu, Mikolov, and Bengio}]{pascanu-etal-2013}
Razvan Pascanu, Tomas Mikolov, and Yoshua Bengio. 2013.
\newblock On the difficulty of training recurrent neural networks.
\newblock {\em ICML (3)\/} 28:1310--1318.

\bibitem[{Saxe et~al.(2013)Saxe, McClelland, and Ganguli}]{saxe2013exact}
Andrew~M Saxe, James~L McClelland, and Surya Ganguli. 2013.
\newblock Exact solutions to the nonlinear dynamics of learning in deep linear
  neural networks.
\newblock {\em arXiv preprint arXiv:1312.6120\/} .

\bibitem[{Seo et~al.(2016)Seo, Kembhavi, Farhadi, and
  Hajishirzi}]{seo-etal-2016}
Minjoon Seo, Aniruddha Kembhavi, Ali Farhadi, and Hananneh Hajishirzi. 2016.
\newblock Bi-directional attention flow for machine comprehension.
\newblock {\em arXiv preprint arXiv:1611.01603\/} .

\bibitem[{Sordoni et~al.(2016)Sordoni, Bachman, and Bengio}]{sordoni-etal-2016}
Alessandro Sordoni, Phillip Bachman, and Yoshua Bengio. 2016.
\newblock Iterative alternating neural attention for machine reading.
\newblock {\em arXiv preprint arXiv:1606.02245\/} .

\bibitem[{Srivastava et~al.(2014)Srivastava, Hinton, Krizhevsky, Sutskever, and
  Salakhutdinov}]{srivastava-etal-2014}
Nitish Srivastava, Geoffrey~E Hinton, Alex Krizhevsky, Ilya Sutskever, and
  Ruslan Salakhutdinov. 2014.
\newblock Dropout: a simple way to prevent neural networks from overfitting.
\newblock {\em Journal of Machine Learning Research\/} 15(1):1929--1958.

\bibitem[{Stolcke(2002)}]{stolcke-2002}
Andreas Stolcke. 2002.
\newblock Srilm --- an extensible language modeling toolkit.
\newblock In {\em Proceedings of the 7th International Conference on Spoken
  Language Processing (ICSLP 2002)\/}. pages 901--904.

\bibitem[{Taylor(1953)}]{taylor-etal-1953}
Wilson~L Taylor. 1953.
\newblock Cloze procedure: a new tool for measuring readability.
\newblock {\em Journalism and Mass Communication Quarterly\/} 30(4):415.

\bibitem[{{Theano Development Team}(2016)}]{theano2016}
{Theano Development Team}. 2016.
\newblock \href{http://arxiv.org/abs/1605.02688}{{Theano: A {Python} framework
  for fast computation of mathematical expressions}}.
\newblock {\em arXiv e-prints\/} abs/1605.02688.
\newblock
  \href{http://arxiv.org/abs/1605.02688}{http://arxiv.org/abs/1605.02688}.

\bibitem[{Trischler et~al.(2016)Trischler, Ye, Yuan, Bachman, Sordoni, and
  Suleman}]{trischler-etal-2016}
Adam Trischler, Zheng Ye, Xingdi Yuan, Philip Bachman, Alessandro Sordoni, and
  Kaheer Suleman. 2016.
\newblock \href{http://aclweb.org/anthology/D16-1013}{Natural language
  comprehension with the epireader}.
\newblock In {\em Proceedings of the 2016 Conference on Empirical Methods in
  Natural Language Processing\/}. Association for Computational Linguistics,
  pages 128--137.
\newblock
  \href{http://aclweb.org/anthology/D16-1013}{http://aclweb.org/anthology/D16-1013}.

\bibitem[{Vinyals et~al.(2015)Vinyals, Fortunato, and
  Jaitly}]{vinyals-etal-2015}
Oriol Vinyals, Meire Fortunato, and Navdeep Jaitly. 2015.
\newblock Pointer networks.
\newblock In {\em Advances in Neural Information Processing Systems\/}. pages
  2692--2700.

\bibitem[{Xiong et~al.(2016)Xiong, Zhong, and Socher}]{xiong-etal-2016}
Caiming Xiong, Victor Zhong, and Richard Socher. 2016.
\newblock Dynamic coattention networks for question answering.
\newblock {\em arXiv preprint arXiv:1611.01604\/} .

\end{thebibliography}
\bibliographystyle{acl_natbib}

\end{document}